\documentclass[manuscript]{acmart}
\AtBeginDocument{%
  }

\setcopyright{acmlicensed}
\copyrightyear{2026}
\acmYear{2026}
\acmDOI{XXXXXXX.XXXXXXX}
\acmConference[Conference acronym 'XX]{Make sure to enter the correct
  conference title from your rights confirmation email}{June 03--05,
  2018}{Woodstock, NY}
\acmISBN{978-1-4503-XXXX-X/2018/06}

\usepackage{graphicx}%
\usepackage{multirow}%
\usepackage{amsmath,amsfonts}%
\usepackage{amsthm}%
\usepackage{mathrsfs}%
\usepackage[title]{appendix}%
\usepackage{xcolor}%
\usepackage{textcomp}%
\usepackage{manyfoot}%
\usepackage{booktabs}%
\usepackage{algorithm}%
\usepackage{algorithmicx}%
\usepackage{algpseudocode}%
\usepackage{listings}%

\usepackage{supertabular,longtable}
\usepackage{pdflscape}
\usepackage{makecell}
\setcellgapes{2.8pt}
\usepackage{verbatim}

\begin{document}

\title{Open Korean Corpora: A Practical Report}


\author{Won Ik Cho}
\affiliation{%
  \institution{AI Center, Samsung Electronics}
  \city{Suwon}
  \country{Korea}}
\email{tsatsuki6@gmail.com}

\author{Sangwhan Moon}
\affiliation{%
  \institution{Google LLC}
  \city{Mountain View}
  \state{California}
  \country{U.S.}}
\email{sangwhan@iki.fi}

\author{Youngsook Song}
\affiliation{%
 \institution{Lablup Inc.}
 \city{Seoul}
 \country{Korea}}
\email{klanguage1004@gmail.com}

\renewcommand{\shortauthors}{Cho et al.}

\begin{abstract}
  Korean is often referred to as a low-resource language in the research community. While this claim is partially true, it is also because the availability of resources is inadequately advertised and curated. This work curates and reviews a list of Korean corpora, first describing institution-level resource development, then further iterate through a list of current open datasets for different types of tasks. We then propose a direction on how open-source dataset construction and releases should be done for less-resourced languages to promote research.
\end{abstract}



\received{xxx}
\received[revised]{yyy}
\received[accepted]{zzz}

\maketitle

\section{Introduction}\label{sec1}

The Korean language is less explored in terms of corpus and computational linguistics, but its prevalence is often underrated. It regards about 80 million language users and is recently adopted in multilingual research as it is bound to CJK (Chinese, Japanese, and Korean), also handling a distinguished writing system.

However, compared to the industrial need, the interest in Korean natural language processing (NLP) has not been developed much in international viewpoints, which recurrently hinders the related publication and further academic extension. Besides, in the recent NLP, where the benchmark practice is a trend, such systems lack at this point, deterring abroad and even native researchers who start Korean NLP from finding directions. \citet{park2016korean} has shown a decent survey, but it seems that the techniques are mainly on the NLP pipeline. Also, albeit some curations on Korean NLP\footnote{\url{https://github.com/datanada/Awesome-Korean-NLP}} and datasets\footnote{\url{https://littlefoxdiary.tistory.com/42}}, we considered that little more organization is required, and better if internationally available. Our attempts are expected to mitigate the challenges that the researchers who handle Korean from a multi- or cross-lingual viewpoint may face.

In this paper, we scrutinize the struggles of government, institutes, industry, and individuals to construct public Korean NLP resources. First, we state how the institutional organizations have tackled the issue by making up the accessible resources, and point out the limitation thereof regarding international availability and license, to finally introduce and curate the fully public datasets along with the proposed criteria. Through this, we want to find out the current state of Korean corpora across the NLP tasks and whether they are freely or conditionally available. Our survey is to be curated and updated in the public repository.
\footnote{\url{https://github.com/ko-nlp/Open-korean-corpora} \\This is Jun. 2026 version (third edition) of the manuscript. You can find the first version in an online archive \cite{cho-etal-2020-open}.}

\section{Accessible Resources}


With the increase in popularity of machine learning-driven methods in NLP, constructing a novel dataset and releasing it to the public can be considered the cornerstone of advancing research of a given language. While we believe many useful datasets exist behind industry walls, this is not particularly useful for advancing open research. Fortunately, there are organizations that construct and distribute cleaned, pre-processed datasets which are occasionally accompanied by a task and the annotation. In the context of Korean, there are numerous efforts in this field driven by government-affiliated organizations.

\subsection{Datasets from public institutions}

\paragraph{Korean Advanced Institute of Science and Technology} (KAIST)  has played a foundational role in the development of Korean computational linguistics. Beginning in the 1990s, research groups such as the Semantic Web Research Center (SWRC), pioneered the creation of essential linguistic resources such as Korean Tree-Tagging Corpus, Morpho-Syntactically Annotated Corpus, Transliteration and Translation Evaluation Sets, Chinese-Korean Multilingual Corpus, etc. These resources have been extensively utilized by researchers worldwide, serving as training data for parsers, benchmarks for shared tasks, and the basis for derivative resources such as the Universal Dependencies  treebank \cite{chun2018building}. However, as institutional web infrastructure has evolved over the years, many of the original distribution links have become deprecated, making direct access to these historically significant datasets difficult for new researchers entering the field.

\paragraph{Linguistic Data Consortium} (LDC) at the University of Pennsylvania has served as a critical infrastructure for distributing standardized Korean language resources to the global research community since its founding in 1992. Through partnerships with various institutions, LDC has curated and maintained an extensive catalog of Korean corpora spanning both text and speech modalities, with key text resources including the Korean Newswire corpus (LDC2000T45, later expanded in LDC2010T19), the Korean Treebank Annotations Version 2.0 (LDC2006T09), the Korean Propbank (LDC2006T03), and most recently, the Penn Korean Universal Dependency Treebank (LDC2023T05). On the speech side, LDC has distributed resources from the CALLFRIEND project (LDC96S54), the Korean Telephone Conversations collection (LDC2003S03/S07), and multilingual collections such as the OGI Multilanguage Corpus and GlobalPhone that include Korean data. Unlike many academic resources with unstable hosting, LDC's institutional model of permanent archiving and standardized licensing has ensured the continued accessibility and citability of these datasets, making them enduring benchmarks for Korean NLP and speech technology research.

\paragraph{National Institute of Korean Language}
(NIKL) is an institution that establishes the norm for Korean linguistics\footnote{\url{https://www.korean.go.kr/}}. However, at the same time, it usually undergoes the massive dataset construction from the view of computational linguistics, to apt to the new wave of language artificial intelligence (AI). Widely known ones include Korean word dictionaries\footnote{The search portal is provided in \url{https://stdict.korean.go.kr/main/main.do} while the full word and content list are available at \url{https://github.com/korean-word-game/db}} and Sejong Corpus \cite{kim2006korean}\footnote{The corpus is officially available through the NIKL  portal at \url{https://kli.korean.go.kr/corpus/main/requestMain.do}.  For preprocessed versions, one can also refer to 
\url{https://github.com/coolengineer/sejong-corpus}.}. The dictionary contains fundamental and new lexicons that make up Korean (along with the content), and the Sejong Corpus is a large-scale labeled NLP pipeline corpus for the tasks such as constituency and dependency parsing, mainly provided in \textit{.xlm}-like format. Besides, recently, labeled corpora of about 300 million word size is released\footnote{\url{https://corpus.korean.go.kr/}}, covering inter-sentence tasks such as similarity and entailment. The corpora (comprising approximately 169 datasets as of February 2026) are continuously being updated regarding typos and inappropriate contents, upon user report and academic feedback.

\paragraph{Electronics and Telecommunications Research Institute}
 (ETRI) has been collecting, refining, and tagging language processing and speech learning data over a long period of time\footnote{\url{https://www.etri.re.kr/intro.html}}. Aside from NIKL, which mainly focuses on classical NLP pipelines, ETRI has also built a database for semantic analysis and question answering (QA), which are the outcome of a project Exo-brain\footnote{\url{http://exobrain.kr/pages/ko/result/outputs.jsp}}. The project includes syntax-semantic ones such as part of speech (POS) tagging and semantic role labeling (SRL),  simultaneously providing construction guidelines for the corpora.

\paragraph{AI HUB}
is a platform organized by 
National Information Society Agency (NIA) in which a large-scale dataset are integrated\footnote{\url{http://www.aihub.or.kr/}}. The datasets are built for various tasks at the government level, to promote the development of the AI industry. Provided resources are labeled or parallel corpora in real-life domains. Here, the domains are law, patent, common sense, open dialog, machine reading comprehension, and machine translation. Also, about 1,000 hours of speech corpus is provided to be used in spoken language modeling\footnote{\url{https://www.aihub.or.kr/aidata/105}}. Recently, some new datasets have been distributed on wellness and emotional dialog, so that many people can have trials for social good and public AI. Also, open dictionary NIADic\footnote{\url{https://kbig.kr/portal/kbig/knowledge/files/bigdata_report.page?bltnNo=10000000016451}} is freely available, provided by K-ICT Big Data Center. As of february 2026, 908 AI training 
datasets are publicly available, of which 184 are Korean language 
datasets, and are continuously managed by the institution.

\subsection{Accessibility}

The above datasets guarantee high quality, along with well-defined guidelines and the well-educated workers. However, their usage is often unfortunately confined to domestic researchers for procedural issues, or when internationally available, not freely open. Researchers abroad (and sometimes domestic as well) can indeed access the data, but they may face difficulty filling out and submitting the particular application form, instead of the barrier-free downloading system. Also, in most cases, modification and redistribution are restricted, making them uncompetitive in view of quality enhancement \cite{han2017open}. 

Here, we want to introduce datasets that can be utilized as an alternative to the limitedly accessible Korean NLP resources. Instead of scrutinizing all available corpora, we are going to curate them under specific criteria.

\section{Open Datasets}

All the datasets to be introduced from now on are fully open access. This means that the dataset is downloadable with a single click or cloning, or at least one can acquire the dataset with simple signing. We set three checklists for the status of the corpus, namely \textit{documentation}, \textit{usage}, and \textit{redistribution}. The first one is on how internationally available the corpus description is. 

\begin{itemize}
	\item Does the corpus have any international article\footnote{Article is here more a complete form of document than a short statistics.} (including paper, blog, Github readme) that incorporates the building process, intended use, etc.? (\textbf{int'l})
	\item Does the corpus have only domestic article available? (\textbf{dom.})
    \item Does the corpus lack an official description? (\textbf{none})
\end{itemize}

Next, we check whether the dataset is both academically and commercially available, academic use only, or unknown \textbf{(all, academic, unknown)}. For the last one, we also investigate if redistribution is available with or without modification, if neither, or unknown \textbf{(rd, rd/mod-x, none, unknown)}\footnote{Though there are various license criteria such as MIT or CC, we did not note them here so as to intuitively display how people should utilize the dataset.}. These attributes are noted along with each corpus title.

We categorize the corpora into ten sections: benchmark studies, parsing and tagging, entailment and similarity, intention understanding and sentiment analysis, offensive language detection and bias, question answering and dialogue, summarization and translation, Korean in multilingual corpora, speech processing, and other specialized domains.

\subsection{Benchmark studies}

Due to emerging PLM studies in Korean and their public release \cite{yang2021transformer}, the need of fair evaluation has grown for a few years. This led to the construction of benchmark dataset that follows GLUE, General Language Understanding Evaluation benchmark \cite{wang2018glue}, and such need had called for the participation of companies and institutions to create new benchmarks that aim to evaluate PLMs' capabilities of Korean understanding. 

The release of ChatGPT in late 2022 marked a paradigm shift in natural language processing, catalyzing significant changes in how Korean NLP resources are developed and evaluated. Prior to this era, benchmark datasets primarily focused on discriminative tasks designed for encoder-based pretrained language models such as BERT \cite{devlin2019bert} and its Korean variants. The post-GPT landscape has witnessed a rapid proliferation of evaluation benchmarks specifically designed for large language models (LLMs), with Korean NLP researchers actively participating in this transition. Benchmarks such as Ko-H5/Open-Ko-LLM, HAE-RAE Bench, KMMLU, and their subsequent refinements (KMMLU-Redux and KMMLU-Pro) exemplify this trend, emphasizing capabilities such as factual knowledge, cultural understanding, and complex reasoning rather than simple pattern matching. 
The emergence of these benchmarks reflects the community's awareness that evaluating generative AI systems requires fundamentally different approaches than those used for discriminative models, including considerations of factuality, cultural appropriateness, and potential harms. Notably, many recent benchmarks incorporate private test sets or contamination detection mechanisms to address data leakage concerns inherent in LLM evaluation.

\paragraph{KLUE (2021)}  \hfill \texttt{[int'l, all, rd]} \\ \citet{park2021klue}\footnote{\url{https://klue-benchmark.com/}} is first non-government-driven Korean language understanding evaluation benchmark created by researchers from multiple organizations (institutes and companies). KLUE consists of eight newly constructed datasets, namely topic classification (KLUE-TC or YNAT, 63K sentences), semantic textual similarity (KLUE-STS, 12K sentence pairs), natural language inference (KLUE-NLI, 31K sentence pairs), named entity recognition (KLUE-NER, 31K sentences), relation extraction (KLUE-RE, 48K sentences), dependency parsing (KLUE-DP, 15K sentences), machine reading comprehension (KLUE-MRC, 29K questions), and dialogue state tracking (KLUE-DST or WoS, 10K dialogues). KLUE aims at NLU benchmark that can evaluate the performance of PLMs not biased to the task specification, text domain or style, or tokenization methodology etc., and is widely used in Korean NLP community as a useful source of training and evaluation.

\paragraph{KoBEST (2022)}  \hfill \texttt{[int'l, all, rd]} \\ \citet{jang2022kobest}\footnote{\url{https://huggingface.co/datasets/skt/kobest_v1}} is a Korean benchmark dataset designed for more challenging language understanding tasks. It comprises five newly constructed datasets: BoolQ (5.8K paragraph-sentence pairs), COPA (4.6K sentence triplets), KB-WiC (5.2K sentence pairs), KB-HellaSwag (3K paragraph and four pairs of sentences), and SentiNeg (4.8K sentence pairs). KoBEST aims to evaluate a model's ability to reason based on more complex knowledge beyond textual form, such as the passage of time, meaning of text, and causality.

\paragraph{Ko-H5/Open-Ko-LLM (2024)} \hfill \texttt{[int'l, all, rd]} \\ 
\citet{park2024open}\footnote{\url{https://huggingface.co/spaces/upstage/open-ko-llm-leaderboard}} is a comprehensive Korean LLM evaluation framework with private test sets to prevent data contamination. Season 1 includes Ko-ARC, Ko-HellaSwag, Ko-MMLU, Ko-TruthfulQA, and Ko-CommonGen v2. Season 2 \cite{kim2025open} adds KorNAT, Ko-GPQA, Ko-WinoGrande, Ko-GSM8K, Ko-EQ-Bench, and Ko-IFEval.

\paragraph{HAE-RAE Bench (2024)} \hfill \texttt{[int'l, all, rd]} \\ 
\citet{son2024hae}\footnote{\url{https://huggingface.co/datasets/HAERAE-HUB/HAE_RAE_BENCH_1.1}} challenges models lacking Korean cultural and contextual depth, covering 6 tasks across vocabulary, history, general knowledge, and reading comprehension domains. About 1,500 questions total.

\paragraph{KMMLU (2024)} \hfill \texttt{[int'l, all, rd]} \\ 
\citet{son2025kmmlu}\footnote{\url{https://huggingface.co/datasets/HAERAE-HUB/KMMLU}} is a comprehensive Korean benchmark with 35,030 expert-level multiple-choice questions across 45 subjects ranging from humanities to STEM. Unlike translated benchmarks, KMMLU is collected from original Korean professional qualification exams, capturing linguistic and cultural aspects unique to Korean.

\paragraph{KULTURE Bench (2024)} \hfill \texttt{ [int'l, academic, unk]}\\ 
\citet{wang2024kulture}\footnote{\url{https://github.com/wangxiaonan-git/KULTUREBench}}  is a benchmark for assessing language models in Korean cultural context, evaluating cultural understanding capabilities specific to Korean society.

\paragraph{KMMLU-Redux/KMMLU-Pro (2025)} \hfill \texttt{[int'l, all, rd/mod-x]} \\ 
\citet{hong-etal-2025-kmmlu} are quality-improved versions addressing critical errors in KMMLU. KMMLU-Redux (2,587 problems)\footnote{\url{https://huggingface.co/datasets/LGAI-EXAONE/KMMLU-Redux}} removes problematic items; KMMLU-Pro (2,822 problems)\footnote{\url{https://huggingface.co/datasets/LGAI-EXAONE/KMMLU-Pro}} comprises 2024 Korean National Professional Licensure exams, manually verified and decontaminated.

\paragraph{KoBALT (2025)} \hfill \texttt{[int'l, academic, rd/mod-x]}\\ 
\citet{shin2025kobalt}\footnote{\url{https://huggingface.co/datasets/snunlp/KoBALT-700}} is a linguistically-motivated benchmark with 700 MCQs spanning 24 phenomena across five linguistic domains: syntax, semantics, pragmatics, phonetics/phonology, and morphology. Expert-curated with human validation by 95 annotators.

\subsection{Parsing and tagging}

As a part of the classical NLP pipeline, we aggregate studies on POS tagging, tree tagging and dependency parsing, named entity recognition (NER) and semantic role labeling (SRL).

\paragraph{UD Korean KAIST (2018)} \hfill    \texttt{[int'l, academic, none]} \\
\citet{chun2018building}\footnote{\url{https://github.com/emorynlp/ud-korean}} applies universal dependency (UD) parsing \cite{mcdonald2013universal} to the Korean Tree-Tagging Corpus \cite{choi1994kaist}. 

\paragraph{PKT-UD (2018)} \hfill \texttt{[int'l, academic, none]} \\
\citet{chun2018building,oh2020analysis}\footnote{Also available at UD-Korean repository, but currently previous version. PKT v2020 data will be uploaded.} applies UD parsing to the Penn Korean Treebank \cite{han2001penn}\footnote{\url{https://catalog.ldc.upenn.edu/LDC2006T09} LDC materials are not curated here.}.

\paragraph{AIR$\times$NAVER NER/SRL (2018)} \hfill   \texttt{ [dom., academic, none]} \\
adopted NER\footnote{\url{http://air.changwon.ac.kr/?page_id=10}} and SRL\footnote{\url{http://air.changwon.ac.kr/?page_id=14}} data constructed by Changwon National University for the purpose of a public competition\footnote{\url{https://github.com/naver/nlp-challenge}}, and  
is annotated according to CoNLL format \cite{tjong2003introduction}. Corpus size is about 90K and 35K each. 

\paragraph{KMOU NER (2019)} \hfill \texttt{[dom., academic, rd]} \\
is an NER dataset built by Korean Marine and Ocean University\footnote{\url{https://github.com/kmounlp/NER}}. The named entities are tagged for about 24K utterances according to name, time, and number. The data source are Exo-brain (by ETRI) and their own data combined, while the redistribution is available only for the latter.


\paragraph{OpenKorPOS (2022)}  \hfill \texttt{[int'l, all, rd]} \\\citet{moon-etal-2022-openkorpos}\footnote{\url{https://github.com/openkorpos}} is a semi-automatically constructed corpus for Korean part-of-speech tagging, built with multiple open-source Korean POS analyzers on Wikipedia dataset. The corpus contains about 55M words (eojeols) and inherits the license of Wikipedia.

\paragraph{KoNEC \& KoNNEC (2022)}  \hfill \texttt{[dom., all, rd]} \\ KoNEC \cite{cheong2022konne}  is an NER dataset\footnote{\url{https://github.com/korean-named-entity/konec}} that annotates 150 types of named entities on the raw corpus of the KLUE-NER data, and KoNNEC\footnote{\url{ https://github.com/korean-named-entity/konne}} is annotated using a nested entity annotation approach on the KoNEC data.

\subsection{Entailment, sentence similarity, and paraphrase}

Here we aggregate corpora for logical inference and checking similarity, as well as style transfer datasets as a part of paraphrase datasets.

\paragraph{Question Pair (2018)} \hfill \texttt{ [dom., all, rd]} \\ consists of about 10,000 open domain sentence pairs\footnote{\url{https://github.com/songys/Question_pair}}, with the binary labels that are hand-annotated on whether the sentences are paraphrase or irrelevant.

\paragraph{Korean PPDB (2019)} \hfill \texttt{[int'l, unk, unk]} \\ 
Korean PPDB (paraphrase database for agglutinative languages) \cite{park2019constructing} is a phrasal paraphrase database specifically designed for agglutinative languages including Korean and Japanese. The authors developed an affix modification-based bilingual pivoting method (AMBPM) to address problems of lexical data sparsity and morphological complexity that standard English PPDB methods cannot handle. The database supports paraphrase generation, question answering systems, and text mining applications. 

\paragraph{KorNLI/KorSTS (2020)} \hfill \texttt{[int'l, all, rd]} \\ 
 \citet{ham2020kornli}\footnote{\url{https://github.com/kakaobrain/KorNLUDatasets}} is a natural language inference (NLI) and sentence textual similarity (STS) dataset for Korean. For KorNLI, the train set was constructed by machine translating SNLI \cite{bowman2015large} and MNLI \cite{williams2017broad}, and the valid and test set were constructed by human translation of XNLI \cite{conneau2018xnli}. Just as in the original dataset, the pairs are labelled with entailment, contradiction, or neutral. About 940K examples are provided for training, and 2,490 and 5,010 respectively for dev and test. For KorSTS, the scoring was done from 0 to 5 to elaborate rather than the binary label that determines paraphrase. Following the scheme of NLI, 5,749 training data were machine translated using the STS-B dataset \cite{cer2017semeval} as a source, while 1,500 dev set and 1,379 test set pairs are human translated.


\paragraph{ParaKQC (2020)} \hfill \texttt{[int'l, all, rd]} \\ \citet{cho2020discourse}\footnote{\url{https://github.com/warnikchow/paraKQC}} originally consists of 10,000 questions and commands, and each instance is labeled with 4 topics (mail, smart agent, scheduling, and weather) and 4 speech acts (\textit{wh-}question, alternative question, prohibition, and requirement). The sentence set can be extended to about 540K sentence pairs that determine sentence similarity and paraphrase.

\paragraph{StyleKQC (2022)} \hfill \texttt{[int'l, all, rd]} \\  
\citet{cho-etal-2022-stylekqc}\footnote{\url{https://github.com/cynthia/stylekqc}} is Korean style transfer and paraphrase dataset that deals with formal and informal Korean questions and commands. It builds upon the construction scheme of ParaKQC and contains 30K sentences, namely 15K for formal and informal style each, built upon 3,000 source phrases and covers six domains regarding smart agents. 

\paragraph{Korean Smile Style Dataset (2022)}  \hfill \texttt{[dom., academic, rd]} \\ 
\citet{SmilegateAI2022KoreanSmileStyleDataset}\footnote{\url{https://github.com/smilegate-ai/korean_smile_style_dataset}}  is colloquial style transfer dataset that contains sentences in 17 styles, built upon a total of about 2,500 dialogs. Styles include formal and informal, robot-like, chat-style, etc., which are casually classified by the dataset builder. 

\paragraph{KoSEnd (2025)}  \hfill \texttt{[int'l, academic, unk]} \\ 
\citet{yu2025making}\footnote{\url{https://github.com/seungukyu/KoSEnd}} is an evaluation dataset grounded in Korean linguistic characteristics, focusing on sentence endings which can significantly change meaning. Collected from three corpora categorized by difficulty (Easy from learner corpus, Intermediate from newspapers, Hard from academic papers).

\subsection{Intention understanding and sentiment analysis}

Beyond corpora that cover sentence pairs, here we introduce some corpora that suit single sentence classification task, that mainly deal with intention or sentiment.

\paragraph{NSMC (2015)} \hfill \texttt{ [int'l, all, rd]} \\   is a review sentiment corpus\footnote{\url{https://github.com/e9t/nsmc}} of size 200K, which consists  of Naver movie comments automatically labeled according to the methodology of \citet{maas-EtAl:2011:ACL-HLT2011}. It adopts pos/neg binary labels, and it has been widely used as a benchmark for pretrained language models. 

\paragraph{3i4K (2018)} \hfill \texttt{ [int'l, all, rd]} \\  
\citet{cho2018speech}\footnote{\url{https://github.com/warnikchow/3i4k}} aims an utterance-level speech act classification of the Korean language. The volume reaches 61K, hand-labeled with 7 classes, namely fragment, statement, question, command, rhetorical question/command, and intonation-dependent utterances.


\paragraph{Kocasm (2019)}  \hfill \texttt{ [doc, all, rd]} \\
\citet{kim2019kocasm}\footnote{\url{https://github.com/SpellOnYou/korean-sarcasm}} is a Korean sarcasm dataset constructed following the collection scheme of \citet{ghosh-veale-2016-fracking}. It contains about 9K Korean tweets crawled online according to some sarcasm-related hashtags and were binary classified manually by authors. 

\paragraph{KMRE (2020)} \hfill \texttt{ [int'l, all, rd]} \\
Korean Movie Review Emotion (KMRE) 
\cite{lee2020korean}\footnote{\url{https://github.com/passing2961/KMRE}} is a large-scale 6-fold emotion-labeled dataset using Korean-specific annotation procedures with n-gram-based distant supervision, built on NSMC.

\paragraph{ToM-Diary (2021)}  \hfill  \texttt{[dom., academic, rd/mod-x]} \\
\citet{lee2021computational}\footnote{\url{https://github.com/humanfactorspsych/covid19-tom-empathy-diary}} is a crowdsourced dataset of 18,238 diaries with 74,014 Korean sentences annotated with Theory of Mind (ToM) levels, designed to measure empathy and perspective-taking ability in written texts. 

\paragraph{KEmoFact (2023)} \hfill \texttt{[int'l, unk, unk]} \\  
\citet{yoo2023korean} is a Korean dataset containing text, emotions, and factors (causes/targets of emotions). Built by translating EmpathicDialogues with manual annotation. Central Supports Emotion Factor Extraction and Emotion-Factor Pair Extraction tasks.

\paragraph{KPC-cF (2024)} \hfill \texttt{ [int'l, academic, rd]} \\  
\citet{nam2024kpc}\footnote{\url{https://anonymous.4open.science/r/KPC-cF-21E8}} consists of Kor-SemEval and KR3. Kor-SemEval is adapted from SemEval-2014 Task 4; KR3 contains actual Korean restaurant reviews with aspect category and polarity labels.

\paragraph{KoCoSa (2024)} \hfill \texttt{ [int'l, all, rd]} \\  
\citet{kim2024kocosa}\footnote{\url{https://github.com/Yu-billie/KoCoSa_sarcasm_detection}} is a dataset for Korean dialogue sarcasm detection consisting of 12,824 daily Korean dialogues (59.3\% sarcasm, 40.7\% non-sarcasm). Created through LLM generation, automatic/manual filtering, and human annotation.

\paragraph{KOTE (2024)} \hfill \texttt{[int'l, academic, rd/mod-x]} \\  
\citet{jeon2024user}\footnote{\url{https://github.com/searle-j/KOTE}} is a large-scale Korean emotion dataset comprising 50,000 online comments with 250,000 annotation cases. Labels 43 emotions plus NO EMOTION through crowdsourcing. Emotion taxonomy derived from cluster analysis of Korean emotion concepts.

\paragraph{CARBD-Ko (2024)} \hfill \texttt{[int'l, academic, unk]} \\  
\citet{jang2024carbd} is a benchmark for aspect-based sentiment classification in Korean with dual-tagged polarities (aspect-specific and aspect-agnostic). Sentences annotated with specific aspects, aspect polarity, aspect-agnostic polarity, and intensity.

\paragraph{KPoEM (2025)} \hfill \texttt{ [int'l, all, rd]} \\
 \citet{lim2025decoding}\footnote{\url{https://github.com/AKS-DHLAB/KPoEM}}
 is a human-labeled dataset for emotion detection in Korean modern poetry, designed for decoding poetic language and emotional expressions.

\subsection{Offensive language detection, fairness and bias}

\paragraph{BEEP! (2020)} \hfill \texttt{ [int'l, all, rd]} \\
\citet{moon2020beep}\footnote{\url{https://github.com/kocohub/korean-hate-speech}} is a hand-labeled, crowd-sourced dataset of about 9.4K Naver entertainment news comments with hate speech and social bias. Bias and hate attribute consists of 3 labels, namely gender/others/none and hate/offensive/none, respectively.

\paragraph{APEACH (2022)} \hfill \texttt{ [int'l, all, rd]} \\
\citet{yang-etal-2022-apeach}\footnote{\url{https://github.com/jason9693/APEACH}} is a balanced evaluation set containing a total of 4K Korean sentences that are either hate speech or non-hate speech. All sentences in the dataset were generated by human participants under the instruction of task managers (authors) and the moderator (of the crowdsourcing platform), given one of ten topics (racism, sexual harassment, gender stereotypes, etc.) per sentence as a condition. 

\paragraph{Korean Unsmile Dataset (2022)}  \hfill  \texttt{[dom., academic, rd/mod-x]} \\
\citet{SmilegateAI2022KoreanUnSmileDataset}\footnote{\url{https://github.com/smilegate-ai/korean_unsmile_dataset}} is an offensive language corpus that contains 19K utterances, namely 10K hate speech, 4K offensive language, and 5K clean expressions. All instances are classified into hate speech, offensive language (comments and profanity terms), and clean expressions, while hate speech can be further annotated with seven multi-label topics. 

\paragraph{HateScore (2022)} \hfill \texttt{ [int'l, academic, rd]}\\ \citet{kang2022korean}\footnote{\url{https://github.com/sgunderscore/hatescore-korean-hate-speech}} is a multilabel hate speech detection corpus that shares the similar construction scheme with Unsmile. It contains 35K instances that consist of 24K online comments, 2.2K neutral sentences from Wikipedia, 1.7K sentences generated human-in-the-loop, and 7.1K rule-generated sentences.

\paragraph{KOLD (2022)}\hfill \texttt{ [int'l, all, rd]}\\ \citet{jeong-etal-2022-kold}\footnote{\url{https://github.com/boychaboy/KOLD}} is Korean offensive language detection corpus that is constructed upon 40K Korean comments from NAVER news and YouTube. Comments are annotated hierarchically with the type (offensive and not-offensive) and the target (untargeted, individual, or group) of offensive language, which also contains the corresponding text spans. The target group and its attribute are also annotated.

\paragraph{K-MHaS (2022)} \hfill \texttt{ [int'l, all, rd]}\\ \citet{lee-etal-2022-k}\footnote{\url{https://github.com/adlnlp/K-MHaS}} is a multi-labeled Korean hate speech dataset built upon 109K utterances from Korean online news comments, tagged in (a) binary manner and (b) 8 fine-grained hate speech classes including politics, origin, physical, age, gender, religion, race, and profanity.

\paragraph{DKTC (2022)} \hfill \texttt{[dom., academic, rd]}\\ \citet{DKTC}\footnote{\url{https://github.com/tunib-ai/DKTC}} is Korean dataset of threatening conversations, which consists of 4K conversations regarding threatening, chantage, bullying, and other harassment (1K each) for train, and 500 test dataset containing conversations without threatening.

\paragraph{KODOLI (2023)} \hfill \texttt{[int'l, all, rd]}\\ \citet{park-etal-2023-feel}\footnote{\url{https://github.com/cardy20/KODOLI}} is a recently published Korean dataset for offensive language identification of size about 38K sentences. It consists of various texts collected and sampled from online communities and news articles, and the texts are tagged in offensive, likely-offensive and none labels. It also contains two auxiliary annotations regarding abusive language and sentiment.

\paragraph{KoMultiText (2023)}  \hfill \texttt{[int'l, all, rd]}\\
 \citet{choi2023komultitext}\footnote{\url{https://github.com/Dasol-Choi/KoMultiText}} is a large-scale Korean multi-task dataset of 150K comments from a Korean SNS platform, annotated for preferences, profanities, and nine types of bias, enabling simultaneous classification of user-generated texts.
 
\paragraph{K-HATERS (2023)} \hfill \texttt{[int'l, all, rd]}\\
\citet{park2023k}\footnote{\url{https://github.com/ssu-humane/K-HATERS}} is the largest Korean offensive language corpus, comprising approximately 192K news comments with target-specific offensiveness ratings on a three-point Likert scale, enabling detection of both explicit and implicit hate expressions.

\paragraph{KoSBi (2023)} \hfill \texttt{[int'l, all, rd]}\\
\citet{lee2023kosbi}\footnote{\url{https://github.com/naver-ai/korean-safety-benchmarks}} is a Korean social bias dataset comprising 34K context-sentence pairs covering 72 demographic groups across 15 categories, designed for filtering-based moderation of biased LLM outputs. 

\paragraph{SQuARe (2023)}  \hfill \texttt{[int'l, all, rd]}\\
\citet{lee2023square}\footnote{\url{https://github.com/naver-ai/korean-safety-benchmarks}}  is a large-scale Korean dataset containing 49K sensitive questions with 42K acceptable and 46K non-acceptable responses, targeting safe response generation for contentious, ethical, and predictive queries. 

\paragraph{KoBBQ (2024)} \hfill \texttt{[int'l, all, rd]}\\ 
\citet{jin2024kobbq}\footnote{\url{https://github.com/naver-ai/KoBBQ}} is a Korean culturally-adapted bias benchmark for QA, with 76,048 samples across 12 bias categories. Includes Korea-specific categories: Domestic Area of Origin, Family Structure, Political Orientation, and Educational Background. Created through large-scale survey validation.

\paragraph{KCDD (2024)} \hfill \texttt{[int'l, academic, none]}\\
 \citet{kim2024towards}\footnote{\url{https://sites.google.com/view/kcdd}} is the first Korean dialogue dataset for violence classification in online settings. Contains 22,249 dialogues with four criminal classes (Serious Threats, Extortion/Blackmail, Harassment in Workplace, Other Harassment) and Clean class.

\paragraph{LifeTox (2024)} \hfill \texttt{[int'l, academic, unk]}\\
\citet{kim2024lifetox}\footnote{\url{https://huggingface.co/datasets/mbkim/LifeTox}} tackles implicit toxicity in life advice, addressing subtle harmful content that may not be detected by standard toxicity classifiers.

\subsection{QA and dialogue}

In this section, we list up QA and dialogue datasets which include question passages or conversations those have significantly longer text compared to sentence-level instances.

\paragraph{KorQuAD 1.0, 2.0 (2019)} \hfill \texttt{[int'l, all, rd/mod-x]}\\ provides human-generated QA corpus and leaderboard for Korean\footnote{\url{https://korquad.github.io/}}. KorQuAD 1.0 \cite{lim2019korquad1} benchmarks SQuAD 1.0 \cite{rajpurkar2016squad} and consists of total 70K questions. KorQuAD 2.0 of size 100K aims at machine reading comprehension for structured HTML natural questions, which was created referring to the scheme of Google Natural Questions \cite{kwiatkowski2019natural}\footnote{\url{https://ai.google.com/research/NaturalQuestions/}}.

\paragraph{KorWikiTQ (2022)}  \hfill \texttt{[int'l, all, rd]}\\
\citet{jun2022korean}\footnote{\url{https://github.com/LG-NLP/KorWikiTableQuestions}} is Korean-specific datasets for table question answering. KorWikiTabular contains tables with descriptions; KorWikiTQ consists of crowdsourced QA pairs with varying difficulty levels.

\paragraph{HuLiC (2022)} \hfill \texttt{ [dom., academic, rd]}\\ consists of human-human conversations of 40K turns and human-machine conversations of 75K turns, where humans talk about movies and human-machine talk about open topics\footnote{\url{https://github.com/smilegate-ai/HuLiC}}. Workers' demographics and other evaluation attributes such as sensibleness, specificity, human-likeness (evaluated turn-wisely), and preference (evaluated every 20 turns).

\paragraph{OPELA (2022)} \hfill \texttt{ [int'l, academic, rd]}\\ \citet{lee2022feels,cho2023crowd}\footnote{\url{https://github.com/smilegate-ai/OPELA}} is a Korean persona dialogue dataset consisting of about 600 conversations, which are created by human participants, namely eleven persona actors (accompanying character profiles) and about 500 user actors. Conversations were made on a chatting app provided by the crowdsourcing platform, where moderators and task managers could monitor the conversation process and moderate the probable issues. Dialogues are additionally annotated with six psychology-related attributes, and 1/3 of the created data was published online for public usage.

\paragraph{CareCall (2022)}   \hfill \texttt{ [int'l, academic, rd]} \\ \citet{bae-etal-2022-building}\footnote{\url{https://github.com/naver-ai/carecall-corpus}} corpus is a Korean "Role specified" Open-domain dialogue in caring senior citizens domain. The dataset is created by using LLMs, along with human support. 10K filtered dialogues are bot-generated via one-shot dialogue generation and human filtering, and each consists of a list of utterances, where each line is tagged with the role (system or user), text, out-of-bounds (boolean whether checks if the utterance of the system violates role specifications). 

\paragraph{CLIcK (2024)} \hfill \texttt{[int'l, all, rd]}\\
\citet{kim2024click}\footnote{\url{https://github.com/rladmstn1714/CLIcK}} is a benchmark comprising 1,995 QA pairs sourced from official Korean exams and textbooks, partitioned into 11 categories across culture (Korean Society, Tradition, Politics, Economy, Law, History, Geography, Pop Culture) and language domains (Textual, Grammatical, Functional Knowledge).

\paragraph{KoDialogBench (2024)}  \hfill \texttt{[int'l, all, rd]}\\ 
\citet{jang2024kodialogbench}\footnote{\url{https://github.com/sb-jang/kodialogbench}} is a benchmark for assessing conversational capabilities of language models in Korean, comprising 21 test sets (82,962 total examples) covering dialogue comprehension (topic, emotion, dialog act classification, fact identification) and response selection tasks.

\paragraph{KorNAT (2024)} \hfill \texttt{ [int'l, academic, rd]}\\ 
\citet{lee2024kornat}\footnote{\url{https://huggingface.co/datasets/datumo/KorNAT}} is the first benchmark measuring alignment between LLMs and South Korean social values. Contains 4K social value questions (validated by survey of 6,174 South Koreans) and 6K common knowledge questions from textbooks/GED exams.

\paragraph{K-MMBench (2024)} \hfill \texttt{ [int'l, academic, rd/mod-x]} \\ 
\citet{ju2024varco}\footnote{\url{https://huggingface.co/datasets/NCSOFT/K-MMBench}}is a Korean adaptation of MMBench for evaluating vision-language models with human-reviewed translations ensuring natural Korean across 4,329 questions with 20 evaluation dimensions.

\paragraph{K-Viscuit (2025)} \hfill \texttt{[int'l, academic, unk]} \\ 
\citet{park2025evaluating}\footnote{\url{https://github.com/ddehun/k-viscuit}}is a multi-choice Visual Question Answering (VQA) dataset for evaluating Vision-Language Models on Korean culture. Created through Human-VLM collaboration focusing on Korean cultural contexts and visual interpretation.

\paragraph{KoSimpleQA (2025)} \hfill \texttt{ [int'l, academic, unk]}\\ 
\citet{ko2025kosimpleqa}\footnote{\url{https://huggingface.co/datasets/bzantium/KoSimpleQA}} is a benchmark for evaluating factuality in LLMs focused on Korean cultural knowledge. Consists of 1,000 short, fact-seeking questions with unambiguous answers, created through crowdsourcing with multiple validation rounds.

\paragraph{KoPIQA (2025)}  \hfill \texttt{ [int'l, academic, unk]}\\ 
\citet{choi2025ko}\footnote{\url{https://huggingface.co/datasets/HAERAE-HUB/Ko-PIQA}} is a Korean physical commonsense reasoning dataset with cultural context. Contains 441 high-quality question-answer pairs, with 19.7\% containing culturally specific elements (kimchi, hanbok, ondol).

\subsection{Summarization, Translation, and Transliteration}

Summarization, translation, and transliteration datasets are separately grouped to specify their usual usage of sequence-to-sequence architectures.

\paragraph{Sci-news-sum-kr (2016)}  \hfill \texttt{[dom., academic, rd]} \\
contains about 50 Korean news summarizations generated by two Korean natives\footnote{\url{https://github.com/theeluwin/sci-news-sum-kr-50}}. Since the size is not large, it is recommended to be used as a dev set.

\paragraph{Korean Parallel Corpora (2016)}  \hfill \texttt{[int'l, academic, rd/mod-x]} \\\citet{park2016korean} contains about 100K en-ko sentence pairs for machine translation (MT). The data mainly bases on news articles, and now also provides the data on North Korean\footnote{\url{https://github.com/jungyeul/korean-parallel-corpora}}.

\paragraph{Transliteration Dataset (2016)}  \hfill \texttt{[dom., all, rd]} \\ is not an official data repository\footnote{\url{https://github.com/muik/transliteration}}, but en-ko transliteration is collected from public dictionaries such as NIKL or Wiktionary\footnote{\url{https://en.wiktionary.org/wiki/Wiktionary:Main_Page}}. A total of about 35K en (word) - ko (pronunciation) pairs are included.

\paragraph{sae4K (2019)} \hfill \texttt{ [int'l, all, rd]} \\\citet{cho2019machines} contains the directive sentence summarization of the sentence level\footnote{\url{https://github.com/warnikchow/sae4k}}. It includes about 50K pairs of utterance and natural language query pair for questions and commands, where the data is partly based on 3i4K \cite{cho2018speech} and some are human-generate in concurrence with \citet{cho2020discourse}. 

\paragraph{OPUS-MT ko-en (2020)}   \hfill \texttt{ [int'l, all, rd]} \\ 
OPUS-MT Korean-English 
\footnote{\url{https://huggingface.co/Helsinki-NLP/opus-mt-ko-en}}
is a part of OPUS project's multilingual translation data \cite{tiedemann2020opus} including Tatoeba Challenge test sets for Korean-English.

\paragraph{Naver News Summary (2022)}  \hfill \texttt{ [none, all, rd]} \\ 
is Korean news summarization dataset crawled from Naver News (IT and economics sections) \footnote{\url{https://huggingface.co/datasets/daekeun-ml/naver-news-summarization-ko}}.

\paragraph{KoreaScience Summary (2023)} \hfill \texttt{ [int'l, all, rd]} \\ 
\citet{nguyen2023loralay}\footnote{\url{https://huggingface.co/datasets/nglaura/koreascience-summarization}} is Korean scientific paper summarization with layout information and structured abstracts.

\paragraph{SSL (2024)}  \hfill \texttt{ [int'l, all, unk]} \\
Korean Sign Language Translation Benchmark \cite{kim2024korean}\footnote{\url{https://github.com/SSL-Sign-Language/Korean-Disaster-Safety-Information-Sign-Language-Translation-Benchmark-Dataset}}is a refined benchmark for Korean sign language translation (disaster safety information). Addresses issues in NIA datasets including computational resources, train/test heterogeneity, and data quality.

\paragraph{KPC (2024)} \hfill \texttt{ [int'l, academic, rd/mod-x]} \\
Korean Unification Parallel Corpus \cite{chun2024bridging}\footnote{\url{https://github.com/HandongSF/KoreanUnificationParallelCorpus}}is a parallel corpus for North Korean-South Korean language translation, containing 130,738 sentence pairs from classic novels (Jane Eyre, The Red and the Black, Korean classics) and the Bible.

\paragraph{KNOTICED (2024)}  \hfill \texttt{[int'l, academic, unk]} \\
\citet{eo2024detecting}\footnote{\url{https://github.com/sugyeonge/KNOTICED}}
is a critical error detection dataset for English-Korean machine translation. Introduces culture-aware "Politeness" error type unique to Korean honorific system. Supports both CED and critical error type classification (CETC) tasks.

\subsection{Korean in multilingual corpora}

We also investigate Korean examples in multilingual corpora, where the type of the dataset may belong to one of the types discussed above.

\paragraph{PAWS-X (2019)}  \hfill \texttt{ [int'l, all, rd]} \\ \citet{yang2019paws} is a dataset that consists of 23,659 human translated PAWS evaluation pairs \cite{zhang2019paws} and about 300K machine-translated ones, for 6 languages including Korean\footnote{\url{https://github.com/google-research-datasets/paws/tree/master/pawsx}}. Among them, Korean occupies about 5K train pairs, and 1,965 and 1,972 for dev/test each. 

\paragraph{Multilingual G2P Conversion (2020)}  \hfill \texttt{ [int'l, all, rd]} \\ \citet{gorman2020sigmorphon} is a shared task of SIGMORPHON 2020\footnote{\url{https://sigmorphon.github.io/sharedtasks/2020/task1/}}, which aims to transform grapheme sequence into a phoneme sequence. The dataset was created with WikiPron\footnote{\url{https://github.com/CUNY-CL/wikipron}} \cite{lee-etal-2020-massively}, and has been built for 10 languages including Korean (3,600 pairs for train, and 450 for dev/test each).

\paragraph{TyDi-QA (2020)}  \hfill \texttt{ [int'l, all, rd]} \\ \citet{clark2020tydi} pursues typological diversity in QA, and provides a total of 200,000 question-answer pairs for 11 linguistically diverse languages, including Korean\footnote{\url{https://github.com/google-research-datasets/tydiqa}}. Among them, Korean occupies about 11K train pairs, and 1,698/1,722 for dev/test each. 

\paragraph{XPersona (2020)}  \hfill \texttt{ [int'l, all, rd]} \\ \citet{lin2020xpersona} is a dataset for evaluating personalized chatbots\footnote{\url{https://github.com/HLTCHKUST/Xpersona}}. It provides the dataset of \citet{zhang2018personalizing} translated to 7 languages, including Korean, where Korean displays 299 dialogues with 4,684 utterances.

\paragraph{XL-Sum (2021)} \hfill \texttt{ [int'l, all, rd]} \\ 
\citet{hasan2021xl} is a large-scale multilingual abstractive summarization from BBC news with professionally written Korean summaries.

\paragraph{MultiCoNER (2022)}  \hfill \texttt{ [int'l, all, rd]} \\ \citet{malmasi-etal-2022-multiconer} is a large multilingual dataset contains 36 NE classes, representing real-world challenges for NER\footnote{\url{https://registry.opendata.aws/multiconer/}}.

\paragraph{MINT (2022)}   \hfill \texttt{ [int'l, unk, unk]} \\  \citet{pei2022semeval} is a multilingual tweet intimacy dataset (MINT) suggested as a task of SemEval 2023\footnote{\url{https://sites.google.com/umich.edu/semeval-2023-tweet-intimacy/home}}. Tweets of six languages labeled in score 1-5 are used in both training and test (each 2K instances), and also tweets of four languages including Korean are used only for zero-shot evaluation (each 500 instances).

\paragraph{MASSIVE (2023)} \hfill \texttt{ [int'l, all, rd]} \\ 
\citet{fitzgerald2023massive} is a 1M-example multilingual NLU dataset spanning 51 languages with virtual assistant utterances for intent classification and slot filling.

\paragraph{IWSLT 2023 (2023)}  \hfill \texttt{ [int'l, all, rd]} \\  holds a formality track that accommodates research in formality translation, where the released dataset consists of 1,000 pairs for en-ko and en-vi each, and 600 zero-shot examples each for en-pt and en-ru\footnote{\url{https://iwslt.org/2023/formality}}. Each pair includes a source English sentence and an informal/formal version of the sentence in the target language. 


\subsection{Speech corpora}

Though most of the datasets discussed in this paper are in text format, we also incorporate speech corpora which can be usefully utilized in speech recognition, speech synthesis, or spoken language processing. 
Speech datasets are usually massive, that a downloading via a single click is not necessarily guaranteed. Thus, we listed some of them as open even if they require some application form. 

\paragraph{KSS (2018)}  \hfill \texttt{ [int'l, academic, rd]} \\\citet{park2018kss} is a book corpus read by a female voice actress. 12K speech utterances and transcriptions are provided\footnote{\url{https://www.kaggle.com/bryanpark/korean-single-speaker-speech-dataset}}.

\paragraph{Zeroth (2018)} \hfill \texttt{ [int'l, all, rd]} \\ is an automatic speech recognition (ASR) dataset that contains approximately 50 hours of well-refined training data\footnote{\url{https://github.com/goodatlas/zeroth}}. The speech corpus is provided free upon request and can be utilized for both research and commercial purposes.

\paragraph{Pansori-TED$\times$KR (2018)}  \hfill \texttt{ [int'l, academic, rd/mod-x]} \\\citet{choi2018pansori} is an ASR dataset obtained by extracting the voices of Korean speakers from Pansori (Korean traditional song in colloquial style) and TED videos, with the transcription augmented\footnote{\url{https://github.com/yc9701/pansori-tedxkr-corpus}}. The total reaches 3 hours, but it incorporates unique phonations that are not viable in other datasets.

\paragraph{ProSem (2019)}  \hfill \texttt{ [int'l, all, rd]} \\  \citet{cho2019prosody} is a spoken language understanding corpus for syntactic ambiguity resolution in Korean, classifying spoken utterances into 7 speech acts\footnote{\url{https://github.com/warnikchow/prosem}}. For about 7,100 utterances recorded by two speakers, namely a male and a female, the ground truth text and label are annotated along with the English translation.

\paragraph{ClovaCall (2020)}   \hfill \texttt{ [int'l, academic, none]} \\ \citet{ha2020clovacall} is an ASR dataset that consists of approximately 80 hours of telephone speech\footnote{\url{https://github.com/clovaai/ClovaCall}}. The corpus is provided upon request, for only research purposes.

\paragraph{JIT/JSS (2020)}   \hfill \texttt{[int'l, all, rd]} \\ \citet{jejueo2020} are datasets containing audio files recorded by a native Jejueo speaker, along with transcript files\footnote{\url{https://github.com/kakaobrain/jejueo}}.


\paragraph{kosp2e (2021)}   \hfill \texttt{ [int'l, academic, rd]} \\  \citet{cho21b_interspeech} is a Korean speech to English text translation dataset that consists of 30K utterances and their Korean script/English translation\footnote{\url{https://github.com/warnikchow/kosp2e}}. The dataset is based on four publicly available corpora, and the license follows each source corpus. It covers various text domains such as news, textbook, AI agent and diary, and all translations were manually performed.

\paragraph{OLKAVS (2024)}  \hfill \texttt{[int'l, all, rd]} \\ 
\citet{park2024olkavs} is the largest publicly available audio-visual speech dataset for Korean. Contains 1,150 hours of audio and 5,750 hours of synced video from 1,107 speakers with nine different viewpoints and various noise conditions.

\paragraph{KMSAV (2024)} \hfill \texttt{ [int'l, academic, rd/mod-x]} \\ 
\citet{park2024kmsav} is an audio-visual speech recognition dataset ($\sim$150h transcribed, 2000+h untranscribed) collected from YouTube videos containing dialogues of multiple participants, designed for spontaneous speech recognition research.

\subsection{Other topics}

Here, we accommodate other Korean datasets that are substantial in their quantity, quality, and documentation, but were not discussed in the previous sections.

\paragraph{K2NLG (2020)}  \hfill \texttt{[dom., academic, rd]} \\  is a dataset for the task of generating summaries from knowledge sets (or knowledge graphs). The object types and relationships of the target knowledge in the model follow the KBox ontology\footnote{\url{https://github.com/machinereading/K2NLG-Dataset}}.

\paragraph{KommonGen (2021)}    \hfill \texttt{ [int'l, all, rd]} \\\citet{jay2021kommongen} is a Korean-generated dataset that translates MS-COCO data captions into Korean and uses them for generalized common sense inference\footnote{\url{https://github.com/nlpai-lab/KommonGen}}. Though only partly available, the team expanded the research to be internationally available \cite{seo-etal-2022-dog} and published the official train set (43K instances) and test set (2K instances) online. The full dataset is available in AI HUB, a government-driven dataset hub for Korean AI, but the access is restricted only to Korean citizens\footnote{\url{https://aihub.or.kr/aihubdata/data/view.do?currMenu=120\&topMenu=100\&aihubDataSe=extrldata\&dataSetSn=459}}.

\paragraph{KoCHET (2022)}    \hfill \texttt{ [int'l, academic, unk]} \\\citet{kim-etal-2022-kochet} is a Korean cultural heritage corpus for Entity-related Tasks, covering the area of named entity recognition, relation extraction (RE), and entity typing (ET), consisting of 112K NER, 39K RE, and 113K ET examples\footnote{\url{https://github.com/Gyeongmin47/KoCHET-A-Korean-Cultural-Heritage-corpus-for-Entity-related-Tasks}}. The construction of the dataset was advised by cultural heritage experts of Korea, and the modified redistribution is allowed for worldwide researchers.

\paragraph{LBox Open (2022)}  \hfill \texttt{ [int'l, academic, rd]} \\ \citet{hwang2022multi} is a multi-task benchmark for Korean legal language understanding and judgement prediction, where the precedent corpus consists of 150K cases of Korean legal precedent (80K from law open data and 70K from the creator's own database)\footnote{\url{https://github.com/lbox-kr/lbox-open}}. Tasks include case name classification (100 classes, 10K pairs), statute classification (46 classes, 2,760 pairs), fine/imprisonment range prediction (10K examples), summarization, etc. 


\paragraph{Korean GEC dataset (2022)} \hfill \texttt{[int'l, academic, rd]} \\ \citet{yoon2022towards} is a secondary-processed dataset for correcting grammatical errors in Korean, derived from the Korean language learner corpus. The Korean GEC dataset can only be used under the same license as the original data\footnote{\url{https://github.com/soyoung97/standard_korean_gec}}.

\paragraph{Korean Ambiguity Dataset (2023)}  \hfill \texttt{[int'l, all, rd]} \\\citet{korean-ambiguity-dataset} is a word sense disambiguation task constructed by Bareun NLP in collaboration with Seoul National University Korean linguistics department, which contains 35K sentences and about 8,200 surface forms of Korean\footnote{\url{https://github.com/bareun-nlp/korean-ambiguity-data}}. It aims at building a comprehensive and objective benchmark on the morpheme-level decomposition of Korean sentences.

\paragraph{KorMedMCQA (2024)}  \hfill \texttt{ [int'l, academic, rd/mod-x]} \\ 
\citet{kweon2024kormedmcqa}\footnote{\url{https://huggingface.co/datasets/sean0042/KorMedMCQA}} is the first Korean Medical Multiple-Choice Question Answering benchmark, derived from professional healthcare licensing examinations (doctor, nurse, pharmacist, dentist) conducted in Korea between 2012-2024. Includes 7,469 questions with Chain-of-Thought annotations by medical professionals.

\paragraph{KBMC (2024)}  \hfill \texttt{ [int'l, academic, unknown]} \\ 
Korean Bio-Medical Corpus \cite{byun2024korean}   is the first open-source medical NER dataset specifically for Korean, created using ChatGPT assistance with human verification, showing 20\% improvement in medical NER performance over previous approaches.

\paragraph{ESG-Kor (2024)}  \hfill \texttt{ [int'l, academic, rd]} \\ 
\citet{lee2024esg} is a dataset for extracting Environmental, Social, and Governance (ESG) information from Korean companies' sustainability reports. Contains 118,946 manually labeled sentences.

\paragraph{KBL (2024)}  \hfill \texttt{ [int'l, academic, rd]} \\ 
\citet{kim2024developing}\footnote{\url{https://github.com/lbox-kr/kbl}} is a pragmatic benchmark for Korean legal language understanding developed with lawyers. Includes 7 legal knowledge tasks (510 examples), 4 legal reasoning tasks (288 examples), and Korean bar exam (4 domains, 53 tasks, 2,510 examples). Also provides Korean precedents corpus (150K precedents).

\paragraph{FunctionChat-Bench (2024)} \hfill \texttt{[int'l, all, rd]} \\ 
\citet{lee2024functionchat}\footnote{\url{https://github.com/kakao/FunctionChat-Bench}} is a novel benchmark for evaluating LLMs' function calling capabilities in Korean dialogues. Contains tool call, answer completion, slot question, and relevance detection annotations.

\paragraph{KCL (2025)}  \hfill \texttt{ [int'l, academic, rd/mod-x]} \\ 
\citet{oh2025korean} is designed to assess language models' legal reasoning capabilities independently of domain-specific knowledge. KCL-MCQA contains 283 questions with 1,103 precedents; KCL-Essay contains 169 questions with 550 precedents and 2,739 rubrics.

\paragraph{KorMedLawQA (2025)} \hfill \texttt{[int'l, academic, rd]} \\ 
is a set of multiple-choice questions focused on South Korean medical law for training LLMs in medical-legal domain and KMLE (Korean Medical Licensing Examination) preparation\footnote{\url{https://huggingface.co/datasets/snuh/KorMedLawQA}}.

\section{Summary}

In total, we surveyed 100 open corpora comprising 82 Korean-specific text corpora, 9 multilingual corpora that include Korean, and 9 speech corpora. By task category, our survey encompasses 8 benchmark studies, 6 datasets for parsing and tagging, 7 for entailment, sentence similarity, and paraphrase, 11 for intention understanding and sentiment analysis, 15 for offensive language detection and bias, 12 for question answering and dialogue, 10 for summarization, translation, and transliteration, 9 Korean components within multilingual corpora, 9 speech corpora, and 13 datasets addressing other specialized domains including legal, medical, and cultural heritage applications. The temporal distribution reveals significant growth since 2018, with peak releases in 2022 (21 corpora) and 2024 (25 corpora), reflecting the Korean NLP community's response to the rise of pretrained language models and subsequent LLM evaluation needs. Regarding accessibility, 53\% of the surveyed corpora permit commercial use, 86\% provide international documentation, and 81\% allow some form of redistribution. We highlight two notable trends: (1) the rapid emergence of Korean-specific benchmark datasets for LLM evaluation, particularly those addressing cultural understanding and factuality, and (2) the substantial growth in offensive language and bias detection resources, reflecting heightened awareness of AI safety concerns in generative model deployment. These trends are expected to continue as LLM development and deployment accelerate. The full specification table of datasets with detailed metadata is maintained and regularly updated in our public repository.

\begin{figure*}[htbp]
    \centering
    \includegraphics[width=\textwidth]{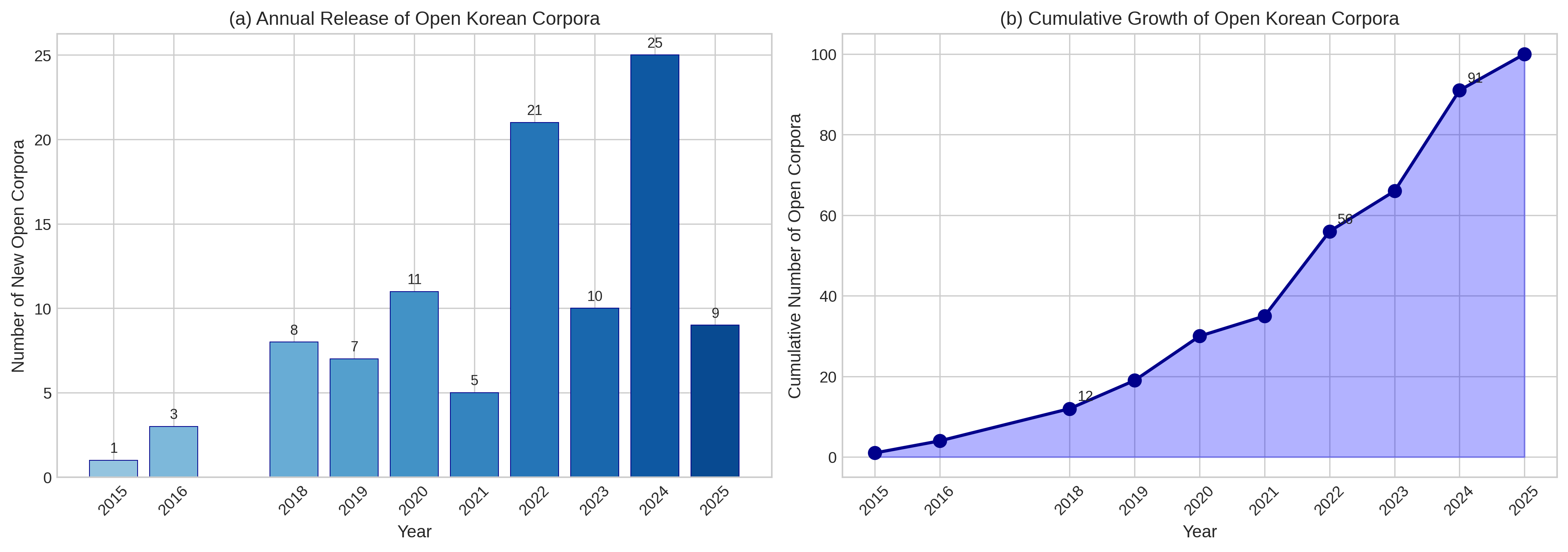}
    \caption{Temporal distribution of open Korean corpora...}
    \label{fig:diachronic}
\end{figure*}

\begin{figure*}[htbp]
    \centering
    \includegraphics[width=\textwidth]{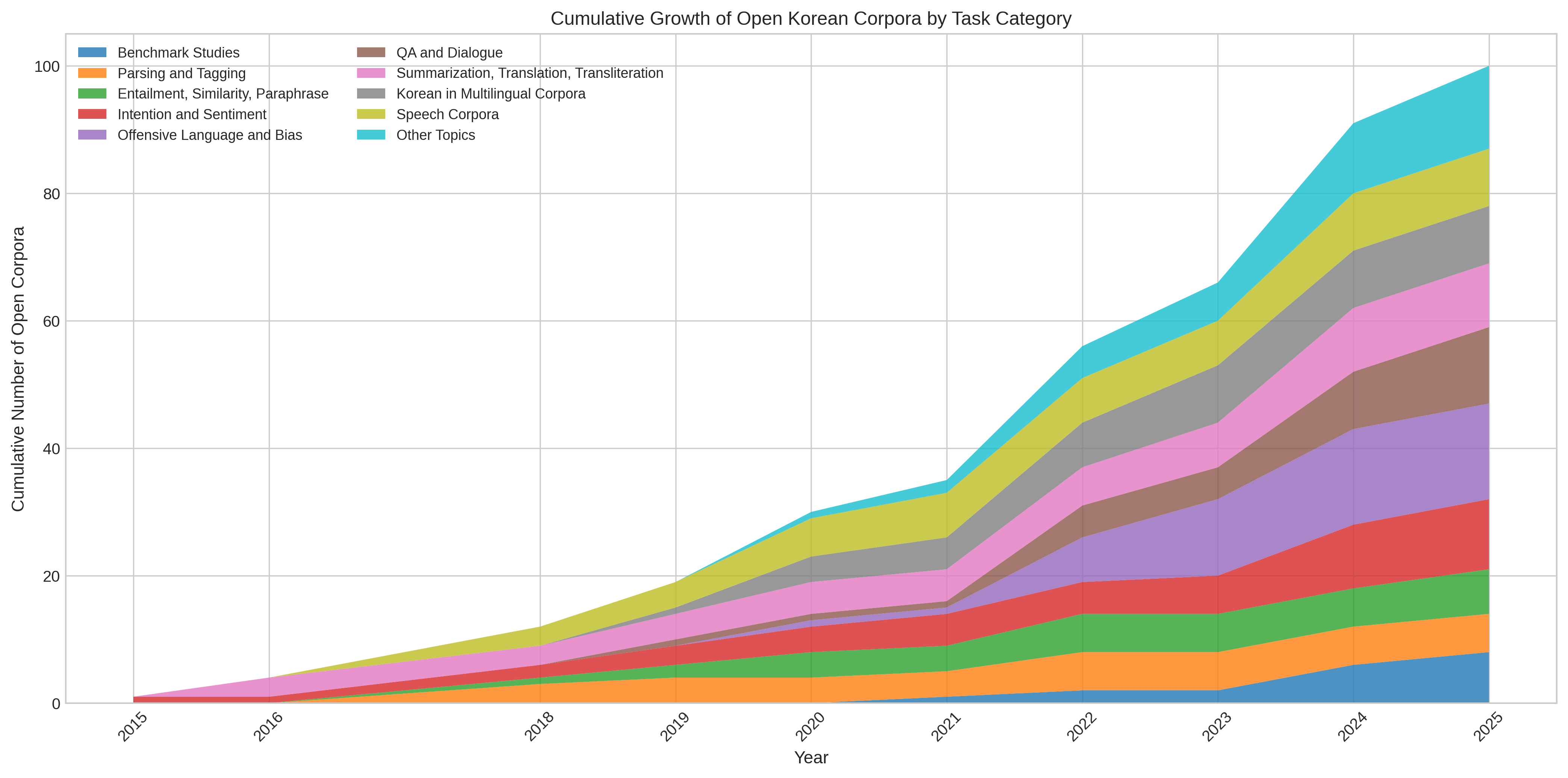}
    \caption{Cumulative growth by task category...}
    \label{fig:by_topic}
\end{figure*}

\subsection{Visualization}
\label{subsec:visualization}

Figure~\ref{fig:diachronic} illustrates the diachronic distribution of open Korean corpora releases from 2015 to 2025. The temporal analysis reveals three distinct phases in the development of Korean NLP resources. The \textit{early phase} (2015--2017) shows minimal activity with only four corpora released, reflecting the period when Korean NLP research was primarily conducted within institutional boundaries with limited open access. The \textit{growth phase} (2018--2021) marks the beginning of substantial community-driven efforts, coinciding with the global rise of pretrained language models such as BERT~\cite{devlin2019bert} and their Korean adaptations~\cite{yang2021transformer}. During this period, foundational resources including parsing corpora (UD Korean KAIST, PKT-UD), sentiment analysis datasets (3i4K, KMRE), and speech corpora (KSS, Zeroth) were established, accumulating to 35 corpora by the end of 2021.

The \textit{acceleration phase} (2022--2025) demonstrates remarkable growth, with 65 new corpora released in just four years. The year 2022 stands out with 21 new releases, driven by the publication of comprehensive benchmark suites such as KoBEST and the KLUE benchmark's widespread adoption, alongside numerous offensive language detection datasets responding to growing concerns about online toxicity. The peak in 2024 (25 releases) reflects the Korean NLP community's rapid response to the large language model era, with the emergence of LLM-specific evaluation benchmarks (Ko-H5/Open-Ko-LLM, HAE-RAE Bench, KMMLU) and culturally-grounded assessment resources (CLIcK, KorNAT). This acceleration aligns with global trends observed in other languages~\cite{liang2023holistic}, where the deployment of generative AI systems has necessitated new evaluation paradigms.

Figure~\ref{fig:by_topic} presents the cumulative growth disaggregated by task category, revealing the evolving priorities of the Korean NLP research community. Several observations merit attention. First, \textit{offensive language and bias detection} has emerged as the largest single category (15 corpora), with the majority released after 2020. This surge corresponds to heightened societal awareness of online hate speech and the need for content moderation in Korean digital platforms~\cite{moon2020beep, jeong-etal-2022-kold}. Second, \textit{benchmark studies} show concentrated growth from 2021 onward, transitioning from discriminative task evaluation (KLUE, KoBEST) to generative model assessment (KMMLU, HAE-RAE Bench), reflecting the paradigm shift from encoder-based to decoder-based language models. Third, \textit{QA and dialogue} resources have expanded substantially since 2022, driven by interest in conversational AI and the need for Korean-specific dialogue evaluation. Fourth, classical NLP pipeline tasks such as \textit{parsing and tagging} show relatively stagnant growth after 2022, suggesting that community efforts have shifted toward higher-level language understanding tasks more relevant to modern LLM applications.

The stacked visualization also highlights the increasing diversification of Korean NLP resources. While early corpora (2015--2018) concentrated on fundamental tasks such as sentiment analysis, parsing, and translation, recent years show parallel development across multiple categories, including specialized domains (legal, medical, cultural heritage) and multimodal applications (audio-visual speech recognition, vision-language benchmarks). This diversification suggests a maturing ecosystem capable of supporting a broader range of research directions and industrial applications.

\begin{table*}[htbp]
\centering
\caption{Summary Statistics of Open Korean Corpora by Category}
\label{tab:summary_by_category}
\begin{tabular}{lcccc}
\toprule
\textbf{Category} & \textbf{Count} & \textbf{Commercial (\%)} & \textbf{Int'l Doc. (\%)} & \textbf{Redistributable (\%)} \\
\midrule
Benchmark Studies & 8 & 75.0 & 100.0 & 87.5 \\
Parsing and Tagging & 6 & 33.3 & 50.0 & 50.0 \\
Entailment, Similarity, Paraphrase & 7 & 57.1 & 71.4 & 71.4 \\
Intention and Sentiment & 11 & 54.5 & 81.8 & 81.8 \\
Offensive Language and Bias & 15 & 66.7 & 86.7 & 86.7 \\
QA and Dialogue & 12 & 33.3 & 91.7 & 75.0 \\
Summarization, Translation, Transliteration & 10 & 60.0 & 70.0 & 80.0 \\
Korean in Multilingual Corpora & 9 & 88.9 & 100.0 & 88.9 \\
Speech Corpora & 9 & 44.4 & 100.0 & 88.9 \\
Other Topics & 13 & 23.1 & 92.3 & 84.6 \\
\midrule
\textbf{Total} & \textbf{100} & \textbf{53.0} & \textbf{86.0} & \textbf{81.0} \\
\bottomrule
\end{tabular}
\end{table*}

\begin{table*}[htbp]
\centering
\caption{Redistribution License Distribution by Category}
\label{tab:redistribution}
\begin{tabular}{lccccc}
\toprule
\textbf{Category} & \textbf{rd} & \textbf{rd/mod-x} & \textbf{none} & \textbf{unknown} & \textbf{Total} \\
\midrule
Benchmark Studies & 5 & 2 & 0 & 1 & 8 \\
Parsing and Tagging & 3 & 0 & 3 & 0 & 6 \\
Entailment, Similarity, Paraphrase & 5 & 0 & 0 & 2 & 7 \\
Intention and Sentiment & 7 & 2 & 0 & 2 & 11 \\
Offensive Language and Bias & 12 & 1 & 1 & 1 & 15 \\
QA and Dialogue & 7 & 2 & 0 & 3 & 12 \\
Summarization, Translation, Transliteration & 6 & 2 & 0 & 2 & 10 \\
Korean in Multilingual Corpora & 8 & 0 & 0 & 1 & 9 \\
Speech Corpora & 6 & 2 & 1 & 0 & 9 \\
Other Topics & 9 & 2 & 0 & 2 & 13 \\
\midrule
\textbf{Total} & \textbf{68} & \textbf{13} & \textbf{5} & \textbf{14} & \textbf{100} \\
\bottomrule
\end{tabular}
\end{table*}

\begin{table}[htbp]
\centering
\caption{Corpus Type Distribution}
\label{tab:corpus_type}
\begin{tabular}{lcc}
\toprule
\textbf{Corpus Type} & \textbf{Count} & \textbf{Percentage (\%)} \\
\midrule
Korean-specific Text Corpora & 82 & 82.0 \\
Multilingual Corpora (incl. Korean) & 9 & 9.0 \\
Speech Corpora & 9 & 9.0 \\
\midrule
\textbf{Total} & \textbf{100} & \textbf{100.0} \\
\bottomrule
\end{tabular}
\end{table}

\subsection{Statistics}
\label{subsec:statistics}

Tables~\ref{tab:summary_by_category}--\ref{tab:corpus_type} present comprehensive statistics on the surveyed corpora, enabling quantitative assessment of the Korean open NLP resource landscape.

\paragraph{Commercial Availability and Usage Restrictions.}
Table~\ref{tab:summary_by_category} reveals that 53\% of the surveyed corpora permit both academic and commercial use, while 44\% restrict usage to academic purposes only. This distribution varies considerably across categories. Multilingual corpora containing Korean data exhibit the highest commercial availability rate (88.9\%), likely because these resources are often developed by international organizations with permissive licensing conventions. Benchmark studies also show high commercial availability (75.0\%), reflecting the community's intent to enable fair model comparison across both academic and industrial contexts. In contrast, domain-specific resources in the ``Other Topics'' category---which includes legal (LBox Open, KBL), medical (KorMedMCQA, KBMC), and cultural heritage (KoCHET) corpora---show the lowest commercial availability (23.1\%), as these specialized datasets often involve licensing constraints from original data sources or institutional policies.

\paragraph{Documentation and International Accessibility.}
A notable strength of the Korean NLP community is the high rate of international documentation: 86\% of corpora provide English-language articles, README files, or technical reports describing their construction methodology and intended use. This represents a significant improvement over the situation described in earlier surveys~\cite{park2016korean}, where many Korean resources lacked accessible documentation for international researchers. The parsing and tagging category shows the lowest international documentation rate (50\%), as several of these corpora were developed for domestic competitions or institutional purposes before the current era of open science practices. All benchmark studies, multilingual corpora, and speech corpora achieve 100\% international documentation, demonstrating the community's awareness that evaluation resources must be globally accessible to enable reproducible research.

\paragraph{Redistribution Licenses.}
Table~\ref{tab:redistribution} details the redistribution policies across categories. Overall, 81\% of corpora allow some form of redistribution, with 68\% permitting full redistribution (including modification) and 13\% allowing redistribution without modification. Only 5\% explicitly prohibit redistribution, concentrated in parsing and tagging datasets (3 corpora) where original licensing from source treebanks imposes constraints. The 14\% with unknown redistribution status highlights an ongoing challenge: even among open datasets, license terms are sometimes ambiguously specified or absent from repository documentation. 

\paragraph{Corpus Type Distribution.}
Table~\ref{tab:corpus_type} summarizes the modality distribution. Korean-specific text corpora constitute the majority (82\%), with multilingual corpora and speech corpora each comprising 9\% of the total. The relatively modest proportion of speech resources reflects both the higher cost of audio data collection and annotation, and the concentration of speech corpus development within government-funded projects (e.g., AI HUB) that may have access restrictions not meeting our open dataset criteria. The equal representation of multilingual and speech corpora (9 each) is coincidental but suggests balanced community attention to cross-lingual research and spoken language processing.

\paragraph{Category-Specific Observations.}
Several category-specific patterns emerge from the statistical analysis. The \textit{offensive language and bias} category shows both high volume (15 corpora) and favorable accessibility metrics (66.7\% commercial, 86.7\% redistributable), indicating strong community commitment to enabling AI safety research. The \textit{QA and dialogue} category, despite its size (12 corpora), shows lower commercial availability (33.3\%) due to the prevalence of academic-only persona dialogue and human evaluation datasets. The \textit{summarization, translation, and transliteration} category demonstrates balanced metrics across all dimensions, benefiting from established machine translation research practices that emphasize reproducibility and open benchmarking~\cite{tiedemann2020opus}.

These statistics collectively indicate that while the Korean NLP community has made substantial progress in open resource development, opportunities remain for improving license clarity, expanding commercial availability in specialized domains, and ensuring comprehensive documentation for legacy resources. We recommend that future corpus construction efforts adopt standardized metadata schemas and explicit licensing from project inception to maximize research utility and international accessibility.

\subsection{Documentation}

Ensuring that a curated list of resources is up-to-date is a challenge. In this regard, we aim to make our work open and canonical, as an online repository of curated resources for Korean. For the research community to have unconstrained access to all current open resources, while endorsing community contributions, the following criteria are crucial:

\begin{itemize}
    \item The canonical, current version of this paper will be regularly published as a revision, e.g., on \url{arxiv.org}, based on a community-open version of this paper.
    \item The resources will also have a corresponding registry, following the same metadata protocol for usability in different types of research, as we used in this protocol.
    \item Each new contribution to the resource list will have a corresponding entry in the acknowledgments section.
\end{itemize}

We will make the registry machine parseable, so that other curated sites such as \url{nlpprogress.com}, can utilize the registry to automate updates. The project will be maintained as an open-source project, under a permissive license. A living document is a new territory for the field of academia, but we strongly believe that given the rapid progress of NLP research, this is an experiment worth attempting; and hope that a successful effort can inspire other languages to follow the same approach. Our approach is to be described in the public repository, guaranteeing the accessibility for domestic and abroad researchers. Also, a large portion of the data are expected to be more easily accessible via open source repositories such as Koco\footnote{\url{https://github.com/inmoonlight/koco}}, Korpora\footnote{\url{https://github.com/ko-nlp/Korpora}}, and hugging face datasets\footnote{\url{https://huggingface.co/datasets}}.

\section{Limitation}

Given recent developments in building Korean PLMs, the need for (commercially) available Korean raw texts has become significant in both industry and open source domains. Though a large portion of government or institution-driven corpora suffices such need, and also Korean web texts have been managed within multilingual web crawl corpora disclosed in venues such as LREC, we have not covered those raw texts in this paper since both types of corpora (namely annotated ones and those for PLM pretraining) differs a lot in terms of the cleanliness/format of the data and the objective of construction. For sure, one may utilize the annotated corpora in their pretraining of PLMs and vice versa -- that is, annotate a part of massive raw text for some specific tasks, but we deemed that the survey on raw texts can be handled more thoroughly within reports of open source (or industry-driven) Korean PLM building projects such as Polyglot \cite{polyglot-ko}, and will possibly redirect the readers to that region for a better opportunity of knowledge.

Another limitation of our study is that, while recent development of large language models such as ChatGPT\footnote{\url{https://openai.com/blog/chatgpt}} has brought a wave of change in data annotation and construction schemes, our study is still focused on datasets that are built in conventional ways -- manual or semi-automatic. We believe that human-generated or human-annotated datasets are still a valuable reference for machine annotation, prompting, or future studies of human behaviors or thoughts, especially in Korean NLP, where the model-centric researches are actively ongoing but computational analysis of language data is still less highlighted. 

We have also deliberately excluded fully synthetic datasets—those generated entirely by language models without substantial human curation—from this survey. While LLM-generated data has emerged as a cost-effective approach for rapid dataset construction, we observe several concerns that warrant careful consideration for Korean NLP research. First, synthetic data may perpetuate biases or errors present in the generating models, which are predominantly trained on English-centric corpora and may not adequately capture Korean linguistic and cultural nuances. Second, the quality and diversity of synthetic Korean text remains difficult to verify at scale, particularly for tasks requiring cultural knowledge or pragmatic understanding. Third, including fully synthetic datasets alongside human-annotated or mediated resources could obscure the actual state of human-curated Korean NLP resources, potentially creating a misleading impression of resource availability. That said, we acknowledge several datasets in our survey that employ hybrid approaches—such as CareCall and KoCoSa—where LLM-generated content undergoes rigorous human filtering and annotation. We consider these human-in-the-loop datasets valid inclusions, as the human validation process addresses quality concerns. As the field matures in understanding how to effectively verify and utilize synthetic data, future revisions of this survey may reconsider this criterion, particularly for datasets with transparent generation methodologies and rigorous quality assurance protocols.

\section{Conclusion}

In this paper, we investigated the Korean NLP datasets constructed and released as public resources. Our curation suggests a variety of open corpora that are freely available. This information will not only be helpful for the Korean researchers who want to start NLP, but also for the abroad ones who are interested in Korean NLP. Nonetheless, we think that Korean open corpora are still less disclosed or not yet sufficient. It is notable that the Korean government is still supplying substantial funds to build a database. To guide this well, appropriate management and documentation should be guaranteed, so that the construction is meaningful and the outcome is internationally available.

\section*{Acknowledgments}

The authors are grateful for all the contributors of the open Korean corpora, including contemporary communities such as EleutherAI,  HAE-RAE, and InstructKR, voluntarily working (or have worked) hard to build and maintain a sustainable ecosystem for Korean NLP. The datasets that this report could not cover will be taken into account in the future revision,  if available. Special thanks go to Seungyoung Lim, Jiyeon Ham, Jiyoon Han, Hyunjoong Kim and Jihyung Moon for checking and proofreading of the earlier versions of the manuscript. We also appreciate team Ko-NLP for accommodating the public repository of our project.

\bibliographystyle{ACM-Reference-Format}
\bibliography{sample-base}










\end{document}